# PICT@DravidianLangTech-ACL2022: Neural Machine Translation On Dravidian Languages

**Aditya Vyawahare** *
aditya.vyawahare07@gmail.com

**Rahul Tangsali** *
rahuul2001@gmail.com

**Aditya Mandke** †
adeetya.m@gmail.com

**Onkar Litake** †
onkarlitake@ieee.org

**Dipali Kadam** ‡
ddkadam@pict.edu

Pune Institute of Computer Technology, India

## Abstract

This paper presents a summary of the findings that we obtained based on the shared task on machine translation of Dravidian languages. We stood first in three of the five sub-tasks which were assigned to us for the main shared task. We carried out neural machine translation for the following five language pairs: Kannada to Tamil, Kannada to Telugu, Kannada to Malayalam, Kannada to Sanskrit, and Kannada to Tulu. The datasets for each of the five language pairs were used to train various translation models, including Seq2Seq models such as LSTM, bidirectional LSTM, Conv2Seq, and training state-of-the-art as transformers from scratch, and fine-tuning already pre-trained models. For some models involving monolingual corpora, we implemented backtranslation as well. These models' accuracy was later tested with a part of the same dataset using BLEU score as an evaluation metric.

## 1 Introduction

Often, it becomes a challenge to develop a robust bilingual machine translation system, and that too with limited resources at hand (Dong et al., 2015). Moreover, for low-resource languages, such as the Dravidian family of languages, achieving high accuracy of translations remains a concern (Chakravarthi et al., 2021). This paper presents the development of machine translation systems for Kannada to other Dravidian languages such as Tamil, Telugu, Malayalam, Tulu, and Sanskrit.

Tamil is a Dravidian classical language used by the Tamil people of South Asia. Tamil is an official language of Tamil Nadu, Sri Lanka, Singapore, and the Union Territory of Puducherry in India (Subalalitha, 2019; Srinivasan and Subalalitha, 2019; Narasimhan et al., 2018). Significant minority speak Tamil in the four other South Indian states of Kerala, Karnataka, Andhra Pradesh, and Telangana, as well as the Union Territory of the Andaman and Nicobar Islands. It is also spoken by the Tamil diaspora, which may be found in Malaysia, Myanmar, South Africa, the United Kingdom, the United States, Canada, Australia, and Mauritius. Tamil is also the native language of Sri Lankan Moors (Sakuntharaj and Mahesan, 2021, 2017, 2016; Thavareesan and Mahesan, 2019, 2020a,b, 2021). Tamil, one of the 22 scheduled languages in the Indian Constitution, was the first to be designated as a classical language of India (Anita and Subalalitha, 2019b,a; Subalalitha and Poovammal, 2018). Malayalam is Tamil's closest significant cousin; the two began splitting during the 13th century AD. Although several variations between Tamil and Malayalam indicate a pre-historic break of the western dialect, the process of separating into a different language, Malayalam, did not occur until the 15th or 17th century (Chakravarthi, 2020; Chakravarthi and Muralidaran, 2021).

One of the approaches implemented consisted of training conventional machine translation models which involved sequence to sequence learning (Seq2Seq) (Sutskever et al., 2014). Seq2Seq is an encoder-decoder approach, in which the encoder reads the input sequence, one word at a time to produce a hidden vector. The decoder produces the output sequence from the vector received from the encoder. We used LSTMs (Hochreiter and Schmidhuber, 1997), Bidirectional LSTMs (BiLSTM) (Clark et al., 2018) which learns bidirectional long-term dependencies between time steps of time series or sequence data, and convolutional Seq2Seq learning (Conv2Seq) (Gehring et al., 2017) which uses multiple stacked layers of CNNs to learn long term dependencies with lower time complexity. The second approach involved training the transformer model (Vaswani et al., 2017) from scratch using the Fairseq library (Ott et al., 2019). We also imple-mented the approach of fine-tuning the open-source translation model provided by AI4Bharat on multilingual data for Indic languages.

---

* equal contribution
† equal contribution
‡ equal contribution

| Parallel | kn-ml | kn-ta | kn-te | kn-tu | kn-sn |
|---|---|---|---|---|---|
| Official | 90,974 | 88,813 | 88,503 | 9,470 | 8,300 |
| **Monolingual** | ml | ta | te | te | te |
| IndicCorp | 80,000 | 80,000 | 80,000 | - | - |

Table 1: Statistics of the dataset used for training

We also fine-tuned their translation models for monolingual data, and then applied back-translation (Edunov et al., 2018; Sennrich et al., 2016a). Back-translation helps avoid the problems caused by the shortage of data for low-resource languages. It is a typical method of data augmentation that can enrich training data with monolingual data. For the ACL 2022 shared task on machine translation in Dravidian languages, we had to submit our results on the five Indic-Indic language pairs: Kannada-Tamil, Kannada-Telugu, Kannada-Malayalam, Kannada-Tulu, and Kannada-Sanskrit. We have experimented and compared the results of the aforementioned models. The datasets were provided by DravidianLangTech. We have used the BLEU (Papineni et al., 2002) evaluation metric for computing accuracy.

## 2 Dataset Description

The bilingual dataset provided by the organizers (Madasamy et al., 2022) was divided into three sub-corporas of train, dev and test. The statistics of the training data is given in Table 1. The dev and test data provided also had the same trend with 2,000 sentence pairs each for Kannada-Malayalam, Kannada-Tamil and Kannada-Telugu whereas Kannada-Sanskrit and Kannada-Tulu had 1,000 sentence pairs each test and dev.

To further improve the accuracy of the translations we used back-translation. The monolingual data used for back-translation was taken from indicCorp (Kakwani et al., 2020) (a large publicly-available corpora for Indian languages created by AI4Bharat from scraping through news, magazines, and books over the web). Monolingual data used was 80,000 each for Malayalam, Tamil and Telugu. We chose 80,000 sentences according to the memory limitations of our GPU. We didn't perform backtranslation on Tulu and Sanskrit as we couldn't find good monolingual data for those languages.

From the monolingual data taken we generate pseudo-parallel data. Using the official and the pseudo-parallel data we train models to provide translations from Kannada to the given Indic languages.

## 3 Data Preparation

In data preprocessing, the sentences present in the given dataset contain punctuations, synonyms, misspelled words, numbers, etc., and they have to be cleaned before we pass it to the model.

For the sentences of Kannada, Malayalam, Tamil and Telugu languages, we used the preprocessing given by indicNLP library [1], which contains preprocessing for various Indian languages. We normalize (helpful in reducing the number of unique tokens present in the text) and then pre-tokenize (for splitting the text object into smaller tokens for better model training) (Harish and Rangan, 2020) the input given followed by transliterating all the indic data written in their own corresponding scripture to Devanagari scripture, along with applying Byte-Pair Encoding (BPE) (Sennrich et al., 2016b). Finally we pass the data to fairseq-preprocess to binarize training data and build vocabularies from the text of that particular language.

For Seq2Seq models such as LSTM and BiLSTM, we took a smaller portion of the dataset, and split it into training data of corpora size 4000, and dev and test datasets of size 1000 for each language pair. For training the Seq2Seq models as well as for training simple transformers from scratch, we used the Sacremoses tokenization [2], where Sacremoses is a pre-installed dependency in the Fairseq toolkit.

## 4 System Description

### 4.1 For Kannada to Malayalam, Tamil, Telugu

In the first system, we download the Indic-Indic model for multilingual neural machine translation given by indicTrans [3] which was trained on the

---
[1] https://github.com/anoopkunchukuttan/indic_nlp_library
[2] https://github.com/alvations/sacremoses
[3] https://indicnlp.ai4bharat.org/indic-trans/

| System | kn-ml | kn-ta | kn-te | kn-tu | kn-sn |
|---|---|---|---|---|---|
| LSTM | **0.3531** | 0.3537 | 0.4292 | 0.5535 | **0.8085** |
| BiLSTM | 0.3352 | 0.3636 | **0.4477** | 0.4200 | 0.8059 |
| Conv2Seq | 0.0233 | 0.0303 | 0.0701 | 0.3975 | 0.4400 |
| Transformer From Scratch | 0.3431 | 0.3496 | 0.4272 | **0.8123** | 0.5551 |
| Pretrained Model | 0.3241 | **0.3778** | 0.4068 | NR | NR |
| Finetuned+Backtranslation | 0.2963 | 0.3536 | 0.3687 | NR | NR |

Table 2: The scores mentioned are the BLEU scores on test data passed. NR represents 'Not Recorded' as the pretrained model did not support translations for those languages. Also, for the LSTM, BiLSTM, and transformer models which were trained from scratch, we used a different test dataset, which was other than the one provided by DravidianLangTech. Results in a similar range would be obtained for the test dataset provided by DravidianLangTech. Highest score achieved for each language pair is marked in bold.

Samanantar dataset (Ramesh et al., 2022). We then generate the pseudo-translations for monolingual data using the same pre-trained transformer_4x multilingual model. Finally, we train the official data and the pseudo-parallel data generated using back-translation to give the translation for the given languages.

The second system which we used was a convolutional neural network (CNN) trained using using the 'fconv' architecture provided by the open-source toolkit fairseq.py. Other Seq2Seq architectures for machine translation included LSTM and BiLSTM, wherein LSTM we construct a standard encoder-decoder LSTM architecture, which is provided in the open-source toolkit fairseq.py

Whereas for BiLSTM we use the same 'lstm' architecture provided, with the only change of making the original encoder parameter as bidirectional. We also trained standard transformer models from scratch, again by using the Fairseq library [4]. Fairseq provides a standard transformer architecture which can be further used for training custom transformer models for machine translation.

### 4.2 For Kannada to Tulu, Sanskrit

In the case of low-resource languages such as Tulu and Sanskrit, there wasn't any support available for multilingual models to be trained on such languages, especially the transformer_4x model, which is a multilingual NMT model by AI4Bharat, trained on the Samanantar dataset (Ramesh et al., 2022). Hence, we were unable to finetune the transformer_4x model and train the multilingual models for these languages as shown in the Table 2 given as Not Recorded (NR). Seq2Seq models (LSTM, BiLSTM, CNN), and transformer models from scratch

---
[4]https://github.com/pytorch/fairseq

were trained. The aformentioned models were trained using the Fairseq toolkit.

## 5 Experiments

### 5.1 Training Details

For training the models we used the fairseq, a sequence model toolkit written in Pytorch (Paszke et al., 2019) developed by Facebook Artificial Intelligence Research (FAIR) team.

We used the custom transformer transformer_4x provided by AI4Bharat and finetuned it on the sum of our official data and pseudo parallel corpora generated. This model was trained with a max-tokens parameter of 1568 and a learning rate of 0.00003 with a label smoothing (Szegedy et al., 2016) of 0.1. For evaluation, we take the best checkpoint from all the checkpoints saved. BLEU was used as the best checkpoint metric and then translations generated were recorded.

We also trained transformer models from scratch which had the architecture consisted of 3 layers each of the encoder and decoder, thus having six stacked layers in the transformer model, The layer size taken was 256 and 3 heads in each attention layer, and the feed forward size for both encoder and decoder was taken to be 512. Each of these transformer models was trained for 10 epochs. The batch size specified during training of these transformer models was 128. Dropout (Srivastava et al., 2014) specified during training was 0.1 . Optimizer used was the Adam optimizer (Kingma and Ba, 2014), and a learning rate of 0.0005. The models were trained on 10 epochs each for every language pair. Using fairseq-generate, we were able to get the BLEU score, which was obtained by comparison between the translated sentences by the model from the source language, with the corresponding target language translations.

For encoder-decoder models involving Seq2Seq learning such as LSTM, BiLSTM and Conv2Seq (using CNNs), we again used the Fairseq toolkit for translation. (reference to the documentation [5]). The LSTM and BiLSTM architectures consisted of a dropout (Srivastava et al., 2014) of 0.2, a learning rate of 0.005, and lr-shrink parameter set to 0.5. Maximum number of tokens in a batch were set to 12000. In case of BiLSTM architectures, the encoder-decoder architecture was made bidirectional. The LSTM and BiLSTM were trained for 25 epochs each. In the case of Conv2Seq, we trained the models for 20 epochs each.

All the above mentioned hyperparameters were giving the best possible results, and hence we proceeded with the use of the same. We finetuned the basic configurations specified in the Fairseq documentation. [6]

## 5.2 Evaluation Metrics

Average sentence BLEU score was used as the evaluation metric. To calculate the BLEU we calculated the score for every sentence and then we averaged the score for the whole corpora of sentences. The BLEU scores were calculated using the sentence_bleu function given by the translate package [7] in NLTK library (Loper and Bird, 2002) with equal weights set to 0.25 for all 4 grams with equal contribution of all 4 grams in the final score. The BLEU scores recorded in Table 2 and Table 4 is scored out of 1. where, closer to 1 means more similarity.

## 6 Results

| Language | Translations |
|---|---|
| kn | ಶೈಕ್ಷಣಿಕ ಅರ್ಹತೆ |
| ml | വിദ്യാഭ്യാസ യോഗ്യതകൾ |
| ta | கல்வித் தகுதி |
| te | అర్ధతలు |
| tu | ಶೈಕ್ಷಣಿಕ ವಿದ್ಯಾರ್ಹತೆ |
| sn | टीकाकारः दक्षता |
| en | educational qualifications |

Table 3: Sample translations taken from the test dataset

---
[5] https://fairseq.readthedocs.io/en/latest/
[6] https://fairseq.readthedocs.io/en/latest/index.html
[7] https://www.nltk.org/api/nltk.translate.html

For the results, please refer to Table 2. The table contains the BLEU scores for the models on which the test data of the language pairs are tested. For the submission of the translations for the language pairs, we used transformer_4x model from AI4Bharat to obtain the translations from Kannada to Tamil, Telugu, and Malayalam. Whereas for the translations from Kannada to Tulu and Sanskrit, transformer models were built from scratch. Results are according to the NLTK BLEU evaluation metric. (After our submission for the workshop task, we explored other models and were getting much better results for the same. You could see those results in the Table 2)

## 7 Competition Results

| kn-ml | kn-ta | kn-te | kn-tu | kn-sn |
|---|---|---|---|---|
| 0.2963 | 0.3536 | 0.3687 | 0.0054 | 0.035 |

Table 4: BLEU scores of the translations submitted to the Machine Translation in Dravidian Languages-ACL2022 shared task

We obtained rank 1 for translations from Kannada to Malayalam, Kannada to Telugu and Kannada to Tamil. For translations from Kannada to Sanskrit and for Kannada to Tulu translations we stood 3rd and 4th respectively (We had initially sent the wrong results for kn-sn and kn-tu for the workshop task submission, hence the low scores were obtained for the same). Results of test sets on the shared task is given in Table 4.

## 8 Related Work

The domain of neural machine translation tasks has been among the interest topics for many researchers. The first machine translation model using deep neural networks was proposed by Kalchbrenner and Blunsom (Kalchbrenner and Blunsom, 2013) . NMT has since been widely studied across the scientific community.

In encoder-decoder mechanisms, the words are converted into word embeddings in the encoder, which are then passed to the decoder which uses an attention mechanism, encoder representations, and previous words to generate the next word in the translation. The encoder and decoder can be deep neural networks such as RNN (Bahdanau et al., 2014), CNN (Gehring et al., 2017), or feed-forward neural networks (Vaswani et al., 2017). Further, there were self-attention models proposed such as transformers which aided to further research in NMT. A notable research related to the efficiency of the same was presented at the proceedings of the 7th Workshop on Asian Translation in 2020 (Dabre and Chakrabarty, 2020). Other related works include those presented at previous ACL conferences in 2019 (Sennrich and Zhang, 2019) and 2020 (Araabi and Monz, 2020).

Pertaining to research in machine translation in Dravidian languages, Xie (Xie, 2021) was able to achieve BLEU scores of 38.86, 36.66, and 19.84 for English-Telugu, English-Tamil, and English-Malayalam using multilingual translation and back-translation. (Koneru et al., 2021) worked on implementing a translation system for English to Kannada by limited use of supplementary data between English and other Dravidian languages. Other works include CVIT's submissions to WAT-2019 (Philip et al., 2019), a transformer-based multilingual Indic-English NMT system (Sen et al., 2018), comparison of different orthographies for machine translation of under-resourced Dravidian languages (Chakravarthi et al., 2019), etc.

## 9 Conclusion

Thus, we implemented neural machine translation systems for Dravidian languages. We utilized different architectures for the same, and analyzed their performance. In future, we plan to train our models with large-scale GPUs. We plan to apply other tokenization methods for the language corpora as well for better training. Also, we plan to train our models with expanded corpora for better results.